\documentclass{article}
\usepackage{spconf,amsmath,graphicx}

\DeclareMathOperator*{\argmin}{argmin}

\usepackage{amssymb}
\usepackage{booktabs}
\usepackage[hang, flushmargin]{footmisc}

\title{CNN in CT Image Segmentation: Beyond Loss Function for \\Exploiting Ground Truth Images\vspace{-6pt}}

\name{Youyi Song$^\star$
      Zhen Yu$^\star$
      Teng Zhou$^\dagger$
      Jeremy Yuen-Chun Teoh$^\ddagger$  Baiying Lei$^\S$
      Kup-Sze Choi$^\star$
      Jing Qin$^\star$\vspace{-6pt}}

\address{$^\star$Center for Smart Health, School of Nursing, The Hong Kong
                 Polytechnic University\\
         $^\dagger$Department of Compute Science, Shantou University\\
         $^\ddagger$ Department of Surgery, The Chinese University of Hong Kong\\
         $^\S$School of Biomedical Engineering, Shenzhen University\vspace{-10pt}}

\begin{document}

\maketitle

\begin{abstract}
  Exploiting more information from ground truth (GT) images now is a new research direction for further improving CNN's performance in CT image segmentation.
  Previous methods focus on devising the loss function for fulfilling such a purpose.
  However, it is rather difficult to devise a general and optimization-friendly loss function.
  We here present a novel and practical method that exploits GT images beyond the loss function.
  Our insight is that feature maps of two CNNs trained respectively on GT and CT images should be similar on some metric space, because they both are used to describe the same objects for the same purpose.
  We hence exploit GT images by enforcing such two CNNs' feature maps to be consistent.
  We assess the proposed method on two data sets, and compare its performance to several competitive methods.
  Extensive experimental results show that the proposed method is effective, outperforming all the compared methods.
\end{abstract}

\begin{keywords}
CT image segmentation, CNN, ground truth image exploitation, network transfer
\end{keywords}

\section{Introduction}
\label{sec:introduction}

Convolutional neural network (CNN), especially U-Net \cite{ronneberger2015u} and its variants, has been proven to be the first choice for segmenting computed tomography (CT) images, a challenging task encountered frequently in clinic practice with a tremendous range of applications, e.g. diseases diagnosis, surgery simulation, therapeutic assistance, and radiotherapy planning, to mention a few \cite{cerrolaza2019computational}.
For training a CNN, ground truth (GT) images play a crucial role, telling the CNN what its output should be and accordingly telling the CNN how to adjust its parameters' value, both by the loss function which aims to measure the expectation of similarity between CNN's output and the GT image.
However, exploiting GT images merely by the loss function often makes the properly trained CNN fail to segment two frequently seen and yet very difficult cases: ($1$) objects to be segmented having similar intensity values to other objects and ($2$) objects having ambiguous borders, mainly because expectation is a rather coarse statistic, unable to offer so rich supervised information for the CNN that these two challenging cases can be well handled.

\vspace{6pt}\noindent\textbf{Related Work:} In order to exploit more information from GT images, there are two types of methods reported.
($1$) \textbf{Regularization}-based methods \cite{oktay2017anatomically,ravishankar2017learning,mirikharaji2018star} focus on devising the loss function.
They model or learn some properties of the objects, and devise those properties as a regularization term.
This class of methods often is possible to obtain a slight performance improvement.
However, it is usually difficult to design a regularization term that is general and optimization-friendly, which substantially weakens the practicability and applicability.
($2$) \textbf{Network Transfer}-based methods \cite{li2018bottleneck,sekuboyina2018btrfly,kakeya20183d}, which are free to design the regularization term, exploit GT images by enforcing the consistency of parameters of two CNNs that are trained respectively on raw CT images and GT images.
Their underlying assumption is that these two CNNs should be similar, because they both are used to segment the same objects.
The behind idea seems to be intuitively correct, but two CNNs might be similar on some unknown metric space, due to the different input spaces.

\vspace{6pt}\noindent\textbf{Contribution:} We present a network transfer-based approach, for exploiting GT images beyond the loss function.
The main technical contribution is a feature similarity module (FSM) that is designed to learn the unknown similarity metric for measuring the similarity of two CNNs trained on raw CT and GT images.
FSM measures the similarity of two CNNs' feature maps rather than the parameters, which seems to be more reasonable because mapping different input spaces into the same output space does need different functions.
Also, FSM no longer requires that two CNNs have the same architecture, increasing the feasibility and practicability.
We assess the proposed method on two CT data sets, and the experimental results show its superiority.

\vspace{-6pt}
\begin{figure*}[!t]
  \centering
  \setlength{\abovecaptionskip}{0pt}
  \setlength{\belowcaptionskip}{0pt}
  \includegraphics [width = 6.8in, height = 1.9in]{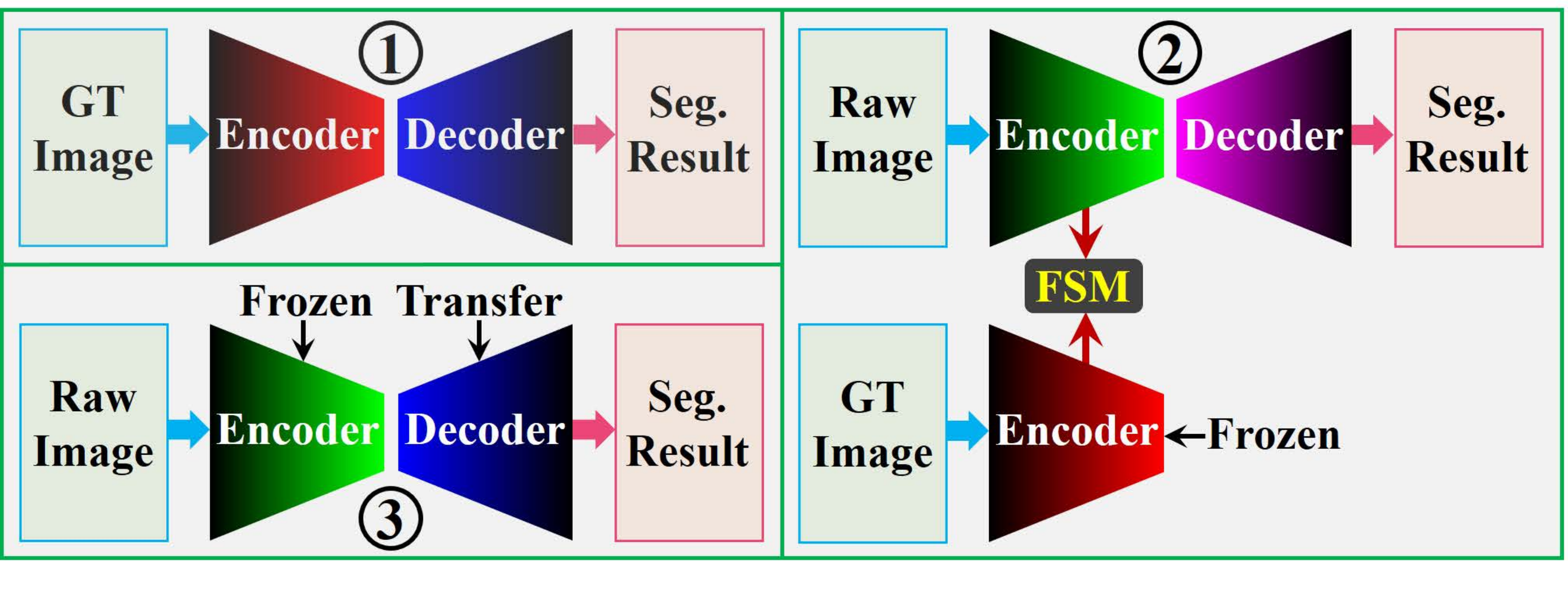}
  \vspace{-10pt}
  \caption{The training strategy for exploiting GT images beyond the loss function.}
  \vspace{-6pt}
  \label{Fig.1}
\end{figure*}

\begin{figure}[!t]
  \centering
  \vspace{6pt}
  \includegraphics [width = 3.0in, height = 1.6in]{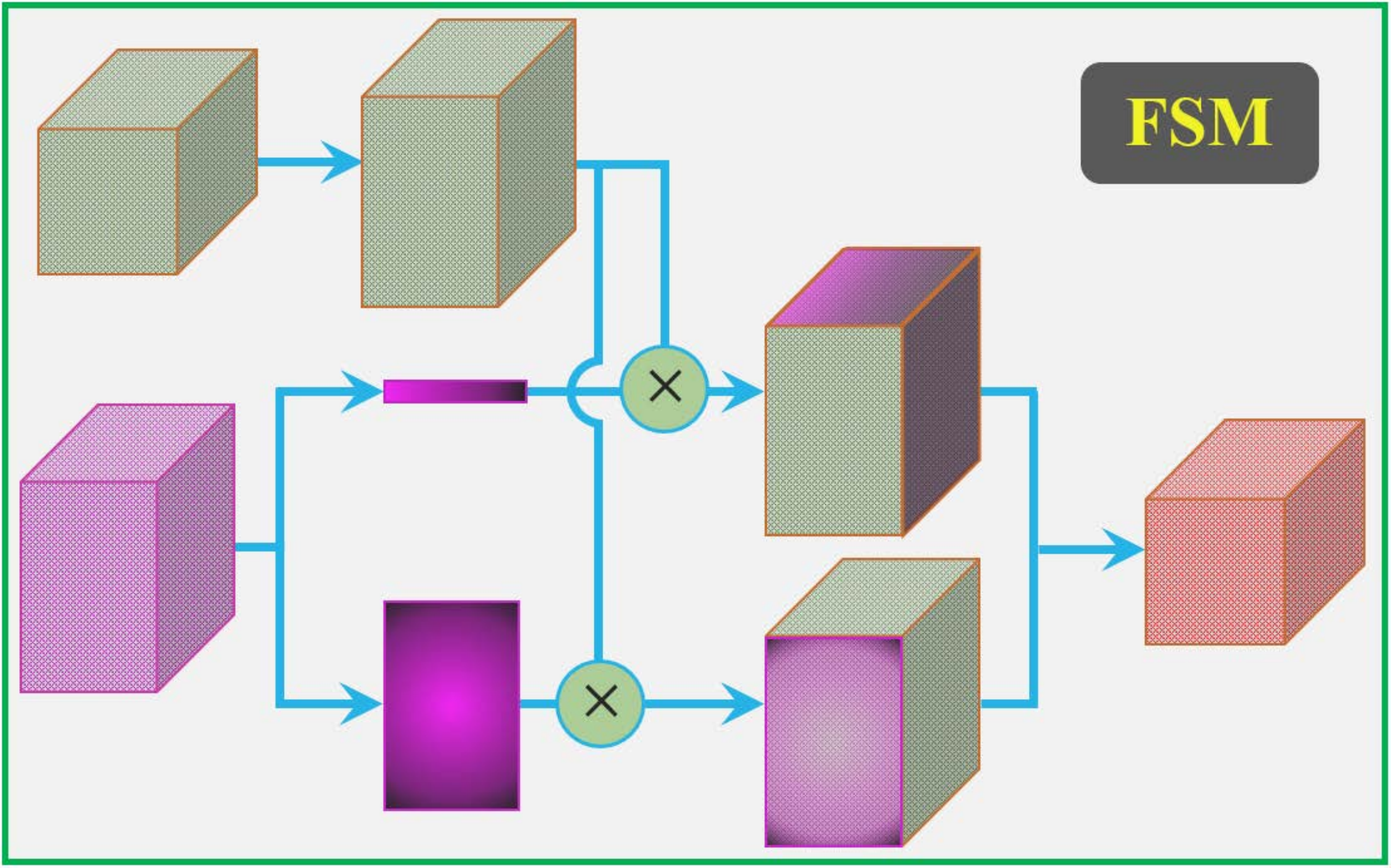}
  \vspace{-8pt}
  \caption{The proposed feature similarity module (without drawing the norm computation part).}
  \label{Fig.2}
  \vspace{-4pt}
\end{figure}

\section{Methodology}
\label{sec:methodology}

In order to exploit GT images beyond loss function, our idea is to transfer knowledge of a CNN, denoted by $\textbf{\emph{N}}_{GT}$, trained on GT images into the CNN, denoted by $\textbf{\emph{N}}_{CT}$, trained for segmenting CT images.
To do so, we need to derive a metric to measure the similarity between $\textbf{\emph{N}}_{GT}$ and $\textbf{\emph{N}}_{CT}$.
We here learn the similarity metric, denoted by $M$, from data, and propose a technique, called feature similarity module (FSM), to fulfill this purpose.
Fig. \ref{Fig.1} shows how we train the CNNs.
We first train $\textbf{\emph{N}}_{GT}$, and then train $\textbf{\emph{N}}_{CT}$ while requiring its feature maps in the encoder part to be similar (measured by $M$) to that of $\textbf{\emph{N}}_{GT}$.
We finally fine tune the decoder of $\textbf{\emph{N}}_{CT}$, which is initialized as the decoder of $\textbf{\emph{N}}_{GT}$.
In what follows, we shall go into technical details.

\subsection{Training Strategy}

We first present our training strategy here, and we then introduce FSM in the next subsection.
Our idea behind training is that feature maps of two CNNs, $\textbf{\emph{N}}_{GT}$ and $\textbf{\emph{N}}_{CT}$ trained respectively on GT and CT images, should be similar on some metric space, because they both are used to describe the same objects for the same purpose.
Intuitively, there are two choices.
The first one is to train two CNNs jointly.
However, the experimental evidence does not show a satisfactory performance improvement (details are presented in the `Experiments' section).
We hence choose the second one, training two CNNs separately, as shown in Fig. \ref{Fig.1}.

More specifically, we first train $\textbf{\emph{N}}_{GT}$ that takes the GT image as the input and tries to output a segmentation result as similar to its input as possible (see the plot $1$ in Fig.~\ref{Fig.1}).
This CNN has the same role as auto-encoder \cite{kingma2013auto}, aiming at learning a high-representative feature of the objects for the segmentation.
Here we employ an U-Net \cite{ronneberger2015u} similar architecture (both $\textbf{\emph{N}}_{GT}$ and $\textbf{\emph{N}}_{GT}$); more implementation details, such as network architecture, loss function, and optimization technique, are presented in the `Experiments' section.

Once $\textbf{\emph{N}}_{GT}$ has been properly trained, we start to train $\textbf{\emph{N}}_{CT}$ (the plot $2$ in Fig. \ref{Fig.1}).
Given a pair of CT and GT images, denoted by $x_i$ and $y_i$, from the training data set $\{x_i, y_i\}_{i=1}^{N}$, $y_i$ is passed through the encoder part of the learned $\textbf{\emph{N}}_{GT}$ to generate feature maps, and $x_i$ is passed through $\textbf{\emph{N}}_{CT}$.
Training then is conducted by enforcing the similarity of two CNNs' feature maps, $F(x_i)$ and $F(y_i)$, in the encoder part, that is,
\begin{align}
\setlength\abovedisplayskip{-10pt}
\setlength\belowdisplayskip{-10pt}
   \{\textbf{w}_{C}^*, \textbf{w}_{M}^*\} = \argmin_{\{\textbf{w}_{C}, \textbf{w}_{M}\}}
   &\sum_{i=1}^N
   \bigg(L\big(f(x_i|\textbf{w}_{C}), y_i\big)\notag\\+
   &\xi M\big(F(x_i|\textbf{w}_{C}),F(y_i)\big|\textbf{w}_{M}\big)\bigg),
\label{Eq. 1}
\end{align}
where $\textbf{w}_{C}$ and $\textbf{w}_{M}$ denote parameters of $\textbf{\emph{N}}_{CT}$ and the similarity metric $M$.
The function $L$ is to measure the similarity between $\textbf{\emph{N}}_{CT}$'s output, $f(x_i|\textbf{w}_{C})$, and the ground truth image $y_i$.
$\xi$ is a balance parameter to control the relative importance of two terms.
The choice of $L$ and the optimal value of $\xi$ are discussed in the `Experiments' section, and details about $M$ will be presented in the next subsection.

We finally transfer decoder's knowledge (the plot $3$ in Fig. \ref{Fig.1}).
We replace the decoder of the learned $\textbf{\emph{N}}_{CT}$ with that of the learned $\textbf{\emph{N}}_{GT}$, and then fine tune it.
We here directly transfer parameters, because features in the encoder part has been enforced to be consistent, so decoders should be similar in the sense of parameters' value.
More implementation details are presented in the `Experiments' section, and below we shall move on to FSM, the proposed feature similarity module.

\subsection{Feature Similarity Module}

FSM aims to measure the similarity of two CNN's feature maps.
Given $F_{CT}^{\ell_i}$ and $F_{GT}^{\ell_j}$, feature maps of $\textbf{\emph{N}}_{CT}$ at layer $\ell_i$ and of $\textbf{\emph{N}}_{GT}$ at layer $\ell_j$ in the encoder part, FSM outputs a scalar between $0$ and $1$ to indicate the similarity, as shown in Fig.~\ref{Fig.2}; the more similar, the larger of the scalar is.
As mentioned before, we allow that two CNNs have different architectures for fully exploiting information from their inputs, so it is possible that $F_{CT}^{\ell_i}$ and $F_{GT}^{\ell_j}$ have different size.
We hence adjust $F_{CT}^{\ell_i}$ has the same size with $F_{GT}^{\ell_j}$, by first adjusting the size of each feature map by nearest interpolating, and then adjusting the channel number by a convolution operation ($3\times 3$ kernel and followed with $ReLU$ \cite{nair2010rectified}) .

As for $F_{GT}^{\ell_j}$, we employ the convolution operations ($3\times 3$ kernel and followed with $ReLU$) to extract its channel and $2$D spatial statistics, denoted by $S_c\in\mathbb{R}^C$ and $S_{s}\in\mathbb{R}^{W\times H}$, where $W$, $H$, and $C$ stand for the width, height, and channel number of $F_{GT}^{\ell_j}$.
Extracted statistics are then used to multiply the adjusted $F_{CT}^{\ell_i}$ (scalar and element-wise multiplications) for transferring channel and spatial knowledge of $F_{GT}^{\ell_j}$ into $F_{CT}^{\ell_i}$.
The multiplied feature maps next are concatenated together and then passed through a convolution operation ($3\times 3$ kernel and followed with $ReLU$) to reduce the channels to that of $F_{CT}^{\ell_i}$.
Finally, we compute Euclidean norm between the resulting feature maps and $F_{CT}^{\ell_i}$ as the similarity value.

\section{Experiments}
\label{sec:experiments}
\vspace{-6pt}
\subsection{Data Sets}

We assess the proposed method on two CT data sets.
The first one is called $3$DIRCADb\footnote{Available on https://www.ircad.fr/research/3d-ircadb-01/}, containing $20$ CT volumes.
Each volume has the same spatial resolutions ranging from $1.6$ to $4.0$ mm.
The second one is called Multi-Organ Abdominal CT\footnote{Available on https://zenodo.org/record/1169361\#.XSFOm-gzYuU}, containing $90$ CT volumes.
Each volume has resolutions ranging from $0.6$ to $0.9$ mm at in-plane and from $0.5$ to $5.0$ mm at inter-slice spacing.

\vspace{-6pt}
\subsection{Evaluation Metrics}

We employ two widely used metrics to evaluate the segmentation performance.
The first one is Dice Similarity Coefficient ($DSC$), aiming at measuring the match degree of the segmentation result and the ground truth.
The second one is Average Symmetric Surface Distance ($ASSD$), aiming at measuring the minimal distance of the segmentation result to the ground truth.
The formal mathematical definition of $DSC$ and $ASSD$ is provided at \cite{yeghiazaryan2015overview}.
Note that a better segmentation algorithm has a larger value of $DSC$ while a smaller value of $ASSD$.

\vspace{-6pt}
\subsection{Implementation Details}

For a fair comparison of the segmentation performance, both $\textbf{\emph{N}}_{GT}$ and $\textbf{\emph{N}}_{CT}$ are chosen as the original U-Net \cite{ronneberger2015u}.
The loss function employed in the training stage $1$ and $3$ ($\textbf{\emph{N}}_{GT}$ training and decoder refine) is $DSC$.
The optimization technique is chosen as Adam \cite{kingma2014adam}, with the initial learning rate as $0.0003$ and the terminated epoch number as $100$ in all three stages.
In addition, in the stage $1$, in order to avoid $\textbf{\emph{N}}_{GT}$ is just to copy the input rather than learning, we put random noise in the input.
Specifically, we randomly set foreground pixels to background pixels with a probability $p$; it is set to $0.2$ by cross validation ($5$-fold).
In the stage $2$, we just enforce the bottom layer, and the balance parameter $\xi$ is set to $0.3$ by cross-validation ($5$-fold).

\vspace{-6pt}
\subsection{Experimental Results}

\begin{table}[t]\small
  \vspace{-8pt}
  \caption{Quantitative Segmentation Results ($DSC$: \% and $ASD$: mm)}
  \vspace{-18pt}
   \begin{center}
       \begin{tabular} {p{0.67in} p{0.38in} p{0.38in} p{0.01in}
                                  p{0.38in} p{0.55in}}\\
           \toprule
           \ &\multicolumn{2}{c}{\ $3$DIRCADb} &\ &\multicolumn{2}{c}{Multi-Organ}\\
           \cline{2-3}
           \cline{5-6}
           \ &\emph{\ \ \ \ DSC} &\emph{\ \ \ ASSD} &\
             &\emph{\ \ \ \ DSC} &\emph{\ \ \ ASSD}\\
           \midrule
           {U-Net}  &{90.4$\pm$2.7} &{1.32$\pm$1.21} &{}
                    &{91.2$\pm$2.4} &{1.25$\pm$1.20}\\

           {U-Net+Loss} &{91.9$\pm$2.1} &{1.17$\pm$1.10} &{}
                        &{92.3$\pm$1.9} &{1.16$\pm$1.06}\\

           {U-Net+Trans} &{92.3$\pm$1.9} &{1.11$\pm$0.99} &{}
                         &{92.7$\pm$1.7} &{1.07$\pm$0.95}\\

           \midrule
            {Joint-Train} &{90.9$\pm$2.5} &{1.27$\pm$1.17} &{}
                          &{91.8$\pm$2.2} &{1.20$\pm$1.14}\\

           {No-Refine} &{93.9$\pm$1.7} &{1.01$\pm$1.01} &{}
                       &{94.2$\pm$1.8} &{0.97$\pm$0.96}\\

           {No-Noise} &{84.2$\pm$6.2} &{1.67$\pm$2.12} &{}
                      &{85.1$\pm$5.7} &{1.51$\pm$1.97}\\
           \midrule
           {U-Net+Our} &{\textbf{94.6$\pm$1.5}} &{\textbf{0.93$\pm$0.91}} &{}
                    &{\textbf{94.8$\pm$1.6}} &{\textbf{0.90$\pm$0.88}}\\

          \bottomrule
       \end{tabular}
   \end{center}
 \vspace{-18pt}
\label{Table 1}
\end{table}

For evaluating the proposed method, we compare its performance to two methods \cite{ravishankar2017learning} and \cite{li2018bottleneck}, denoted respectively by U-Net+Loss and U-Net+Trans.
U-Net+Loss exploits GT image by devising the loss function, while U-Net+Trans by transfer networks' parameters.
For a fair comparison, all methods employ the same network (U-Net) and the same learning setting (the optimization technique and terminated epoch number), while hyper-parameters' value are chosen according to authors' recommendation.
Results reported here are on liver, left kidney, and right kidney, by a $5$-fold cross validation.

\vspace{6pt}\noindent\textbf{Quantitative Results:} We first look at quantitative results, provided in Table.~\ref{Table 1}.
By following the conventional way, we calculate $DSC$ to three decimal places and report in a percentage manner, while $ASD$ to two decimal places and report its real number.
From Table.~\ref{Table 1}, we can see that all three methods can improve the segmentation performance, and that the proposed method improves the most ($4.2\%$ on $DSC$ and $0.37$ mm on $ASSD$ on average), demonstrating the effectiveness of the proposed method.

\begin{figure*}[!t]
  \centering
  \includegraphics [width = 7in, height = 1.9in]{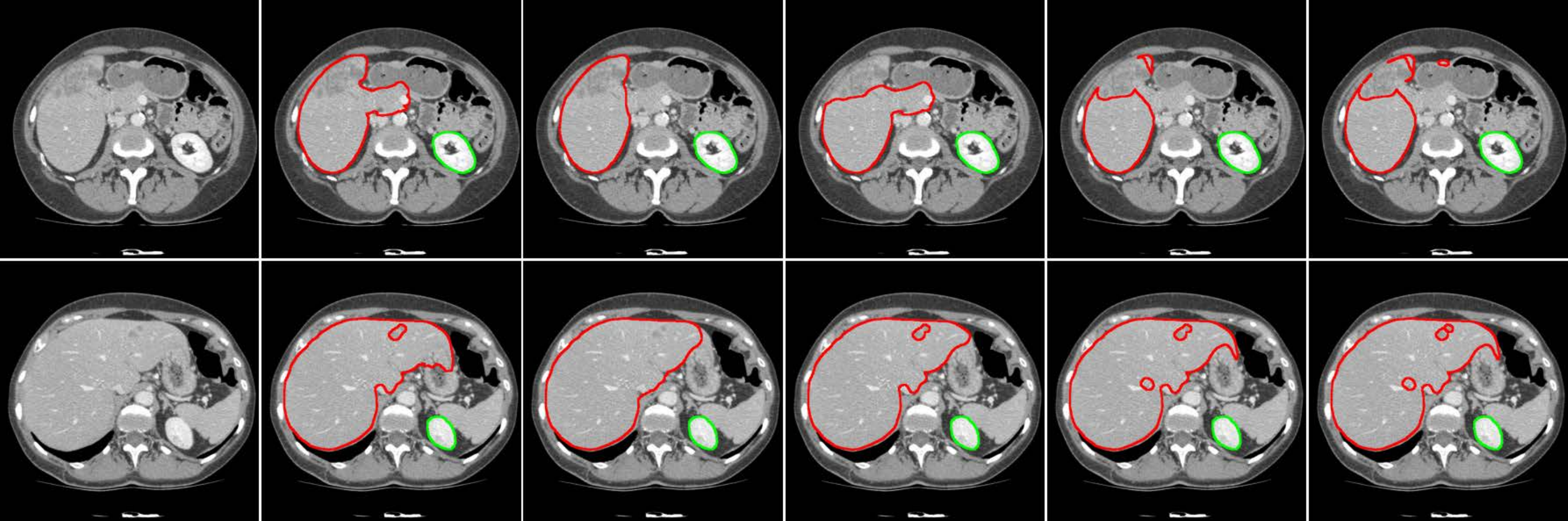}
  \vspace{-16pt}
  \caption{Two visual examples in which the liver has ambiguous border and some of its region has the similar intensity value to some background region. From left to right columns: CT image, results produced by U-Net \cite{ronneberger2015u}, U-Net+Loss \cite{ravishankar2017learning}, U-Net+Trans \cite{li2018bottleneck}, and the proposed method, and GT image. Liver is depicted by red color while left kidney by green color; right kidney does not appear here.}
  \vspace{-10pt}
  \label{Fig. 3}
\end{figure*}

\vspace{6pt}\noindent\textbf{Qualitative Results:} We now look at two visual examples, as shown in Fig.~\ref{Fig. 3}, for qualitatively evaluating the performance improvement of the proposed method.
In the two examples, the liver has ambiguous border and some of its region has the similar intensity value to some background region.
We can see from Fig.~\ref{Fig. 3} that results produced by the proposed method are the most similar to GT images.
Specifically, the original U-Net fails on both cases.
U-Net+Loss can distinguish those intensity-similar pixels, however it cannot distinguish the ambiguous border.
U-Net+Trans cannot consistently distinguish those intensity-similar pixels, and also fails to distinguish the ambiguous border.
The proposed method seems to be the only one that can deal with these two difficult cases, demonstrating its effectiveness again.

\subsection{Ablation Study}

We here aim to demonstrate the importance of each component of the proposed method on the performance improvement.
We did three ablation experiments.
The first one, denoted by Joint-Train, trains two CNNs jointly, for demonstrating the training strategy.
The second one, denoted by No-Refine, removes the training stage three, without fine tuning the decoder.
The third one, denoted by No-Noise, does not put random noise in the input in the training stage one.
Results are reported in Table.~\ref{Table 1}.
We can see that all three components are necessary and mutually reinforcing.

\vspace{-2pt}
\section{Conclusion}
\label{sec:conclusion}

For further improving CNN's performance in CT image segmentation, a challenging and yet frequently encountered task in clinical practice, in this paper we investigate the importance of enforcing the similarity of feature maps of two CNNs trained respectively on CT and GT images.
Compared to two types of existing methods, regularization-based and network transfer-based, this idea is firmly verified on two CT data sets and the experimental results show that it is more effective.

\vspace{-6pt}
\section*{Acknowledgement}
\vspace{-6pt}

The work described in this paper is supported by a grant from the Hong Kong Research Grants Council (No. PolyU $152035$/$17$E).
\vspace{-6pt}
\bibliographystyle{IEEEbib}
\bibliography{refs}

\end{document}